\documentclass[letterpaper]{article}

\usepackage{amsmath}
\usepackage{natbib,alifeconf}  
\usepackage{url,hyperref,cleveref}
\usepackage{booktabs}
\usepackage{comment}
\usepackage{xcolor}

\hypersetup{
           breaklinks=true,   
           colorlinks=true,   
           pdfusetitle=true,  
        }
\title{ Meta-Learning an Evolvable Developmental Encoding }

\author{
    Milton L. Montero$^*$,
    Erwan Plantec,
    Eleni Nisioti,
    Joachim W. Pedersen, \and 
    Sebastian Risi$^*$ \\
    \mbox{}\\
    IT University of Copenhagen, Denmark \\
    $^*$\{mlle,sebr\}@itu.dk
} 

\begin{document}

\maketitle


\begin{abstract}

Representations for black-box optimization methods (such as evolutionary algorithms) are traditionally constructed using a delicate manual process. This is in contrast to the representation that maps DNAs to phenotypes in biological organisms, which is at the heart of biological complexity and evolvability. Additionally, the core of this process is fundamentally the same across nearly all forms of life, reflecting their shared evolutionary origin. Generative models have shown promise in being learnable representations for black-box optimization but they are not per se designed to be easily searchable. Here we present a system that can meta-learn such representation by directly optimizing for a representation’s ability to generate quality diversity. In more detail, we show our meta-learning approach can find one Neural Cellular Automata, in which cells can attend to different parts of a ``DNA’’ string genome during development, enabling it to grow different solvable 2D maze structures. We show that the evolved genotype-to-phenotype mappings become more and more evolvable, not only resulting in a faster search but also increasing the quality and diversity of grown artefacts. 
\end{abstract}

\section{Introduction}

Evolutionary computation algorithms used for black-box optimization have two fundamental requirements: 1) determining the data structure that describes valid solutions and 2) the variation operations that modify said solutions during optimisation \citep{rothlauf2006representations,ashlock2012representation}. For complex problems, a well-chosen representation can significantly enhance the effectiveness of the particular operators being used. Researchers often imposing properties such as symmetric and repeating structures \citep{stanley2007compositional}, or modularity \citep{doncieux2004evolving,schrum2014evolving}. However, crafting the ``right'' representations on  case-by-case basis is a challenging problem, since it requires identifying the characteristics of the design space over which such  properties could be imposed. Optimal representations strike a balance between emphasizing high-quality solutions, making them readily accessible, and supporting a diversity of potential solutions, allowing for extensive exploration of the search space. 

Instead of designing representations by hand, they can also be learned when a dataset of high-quality solutions is available. More and more powerful generative models have shown that this approach can create a large diversity of artefacts.
Recently, generative models have been combined with evolutionary approaches, by using them to search through the learned latent spaces of Generative Adversarial Networks (GANs) and Variational Auto-encoders (VAEs) \citep{volz2018evolving, bontrager2018deepmasterprints}. To extend the expressivity of these models beyond what was part of the original training data, they can also be trained further on high-quality solutions discovered through evolution \citep{gaier2020discovering,moreno2018learning} However, while these representations generate diverse and high-quality solutions, they are \emph{not directly optimized for being easily searchable}. A representation might include the artefacts we are looking for but this property is only really useful if the employed optimization method is able to navigate the fitness landscape induced by that representation \citep{volz2023tools,ashlock2012representation}. 

The main insight in this paper is that it is possible to learn a representation that is easily searchable by a particular black-box optimization algorithm, i.e. it can be used to generate a diversity of high-quality solutions using an evolutionary algorithm. We achieve this by using a novel meta-learning approach in combination with a variant of a developmental encoding, namely a Neural Cellular Automata (NCA) \citep{mordvintsev2020growing,palm2022variational,earle2022illuminating,sudhakaran2021growing,niklasson2021self}. NCAs have been shown to be very expressive, being able to encode a large diversity of different artefacts with interesting regularities without the need to specifically bias the representations towards them. In contrast to previous NCA work, the particular NCA representation in this paper can be conditioned on a ``DNA'' input string, to which each cell can attend during growth, allowing one NCA to produce a diversity of artefacts (Figure~\ref{fig:model-and-training}a). In other words, we separate the genotype-to-phenotype mapping (NCA) from the genomic input (DNA). To find one NCA that can generate a diversity of high-quality solutions, we evaluate its ability to perform a quality-diversity (QD) search \citep{lehman2011evolving,mouret2015illuminating}:  For each NCA, we run MAP-Elites in the inner loop, evolving a population of DNA string to maximize that NCA's QD score (Figure~\ref{fig:model-and-training}b).

We compare this approach with an alternative model that uses a continuous embedding as a genotype and show that it is superior on a simple level generation task. Additionally, we analyze the properties of the system, specifically how the DNA effectively guides development. We also show that the genotype-to-phenotype mapping becomes more evolvable over time, reducing the time it takes to find high-quality and diverse solutions. Finally, we study the properties of the evolved DNA genomes using a variety of different measures such as total disentanglement and information gap.  

\section{Background}

\subsection{Evolvability}
``Evolvability'' is a concept in evolutionary biology that refers to the capacity of a biological system — such as a population, an organism, or a genetic architecture — to generate heritable phenotypic variation that can be acted upon by natural selection. In simpler terms, it describes how capable a system is of evolving. In this paper, we define evolvability as the capacity of an organism to ``generate heritable phenotypic variation'' \citep{kirschner1998evolvability}.

\citet{lehman2016critical} showed that there is a fundamental connection between a more divergent selection process and evovability, both on the individual and population level. In the work presented here, we specifically optimize for evolvability by rewarding individual NCAs for a high QD-score.

Alternatively, it is possible to directly optimize for Evolvability. In an approach (conveniently) called Evolvability Search \citep{henok2016}, the potential for future diversity in an individual is measured by the behavioral diversity of its direct offspring. Organisms that exhibit greater variation among their offspring are then directly optimized for. Follow up work, called Evovability ES, showed that it is also possible to directly reward representations to be highly evolvable without the need for this computational intensive estimation by optimizing an evolvability-inspired objective \citep{gajewski2019evolvability}. In general, however, these approaches optimize for diversity and do not take quality into account (beyond measuring viability).

\subsection{Learning Representations}
Learning representations has a long history in evolutionary computation but is still not very well understood \citep{scott2015learning,bongard2003evolving,simoes2014self,altenberg1994evolving}. Particularly relevant to our approach is the work by \citet{simoes2014self} in which the authors evolve both a genome and the genotype-to-phenotype mapping. However, in contrast to our approach, the mapping is not a developmental encoding and not optimized to create a large diversity of high-quality solutions. 

Supervised methods have also been successfully employed to learn representations. Generative models such as GANs \citep{goodfellow2020generative}) and VAEs \citep{kingma2013auto} have allowed representations to be directly learned from example data. Approaches that employ evolutionary algorithms to search inside these learned latent spaces for artefacts with specific properties, also called latent variable evolution (LVE) approaches \citep{bontrager2018deepmasterprints}, have been applied to a variety of different domains \citep{volz2018evolving, bontrager2018deepmasterprints, olesen2021evolutionary}. 

If no data set with high-quality solutions is available for a supervised learning approach, generative models can be repeatedly trained through a bootstrapping approach on discovered high-quality solutions  \citep{moreno2018learning, gaier2020discovering, torrado2020bootstrapping}. 

\subsection{Neural Cellular Automata}
Neural Cellular Automata (NCA), replace the traditional lookup table in Cellular Automata with a learned neural network. Importantly for this paper, they have been shown to be a very expressive encoding, able to grow complex 2D image \citep{mordvintsev2020growing}, video game levels \citep{earle2022illuminating}, and even 3D Minecraft structures \citep{sudhakaran2021growing}.

\begin{figure*}[t!]
    \centering
    \includegraphics[width=\textwidth]{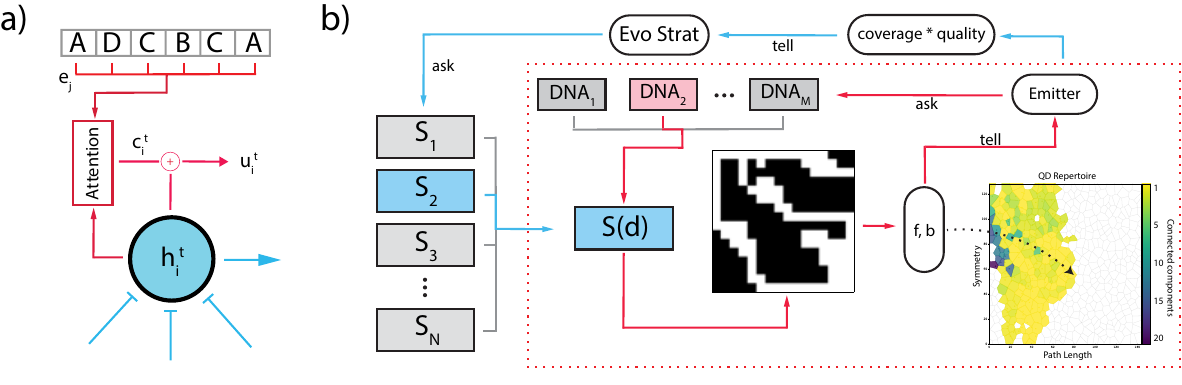}
    \caption{\textbf{Meta-learning DNA-like representations}. (\textbf{a}) The sketch of DNA-guided development. A component $i$ of a developmental system with some hidden state $\mathbf{h}_i^t$ sends and receives information from other components in the system (blue arrows). Additionally, the DNA encoding (gray) provides a global set of instructions via state-dependent decoding mechanism (red). The decoding mechanism produces an output $\mathbf{c}_i^t$ which conditions the state vector (plus sign) producing a vector $\mathbf{u}_i^t$ which is then used for any downstream operation in the node (e.g.\ determining which information to send to other components). (\textbf{b}) General procedure to train a parameterized developmental system $S$ (e.g.\ NCA). An outer loop evolutionary strategy generates a population of possible parameters for $S$. Each of these (blue) is evaluated (inside the red square) on their capacity to fill an archive of solutions (in this example, simple mazes) defined according to their quality and diversity values. The developmental system must achieve this by maximizing its usage of the potential structure in the DNA-like encodings (red).}
    \label{fig:model-and-training}
    \vspace{-0.3cm}
\end{figure*}

Additionally, these systems were shown to be robust to corruptions introduced during the generative process. However, the standard NCA approach is limited in the fact that each NCA only grows one particular pattern, thereby limiting their usability as a generative model.

Several related works have attempted to overcome this limitation in different ways. \citet{sudhakaran2021growing} extended the basic NCA model to use one-hot vectors as inputs. Each of these vectors would correspond to a different class in the emoji dataset introduced in \citet{mordvintsev2020growing}, and can be used to modify the NCA states in a class-dependent way during development. Thus the NCA could be trained to generate more than output in accordance with the target vector.

On the other hand, \citet{palm2022variational} showed how the NCA approach could be combined with the autoencoder architecture to generate a distribution of examples without class supervision, though it does require a dataset of examples for training, much like any other auto-encoder variant.

A different approach introduced in \citet{earle2022illuminating} is to forgo any attempt to learn an NCA that can generate a large diversity of artefacts  and instead learn a population of NCAs, each of which generates a different instance. To ensure that the NCA population is not limited to known classes they use quality-diversity training, defining a feature space of interest that the training procedure must cover using the different NCA instances (their model was applied to simple 2D level generation).

\section{Methods}

We now describe both the general architecture of our model and its training method. Based on the properties observed in biological developmental systems, the proposed encoding should exhibit the following three properties: (\textbf{1}) The genotype-to-phentoype mapping should be able to generate diverse entities when conditioned on different DNA genomes. Additionally, each cell should be able to attend to different parts of the DNA-string during the growth process, allowing the DNA to guide this self-organizing process. 

    (\textbf{2}) The latent space described by such as mapping should possess structure, preferably one that is easy to manipulate/explore. 
     (\textbf{3}) Ideally, the system should not rely on supervised training. 
     That is, it should at most receive some guidance regarding what are the properties of interest for a specific domain, but should not require specific examples.

It is easy to satisfy the first requirement by conceptualizing the generative process as a self-organizing system unfolding through time. The DNA genome can then bias state updates at each time step. 

The second property can be satisfied by taking inspiration from DNA itself and parameterize it as a string of characters. Notice that this is similar to the LLM approach used in \citet{Nasir2023LLMaticNA} except that we \emph{let the model define it's own language} based on how the encoding is used by the generative process. This could potentially overcome a limitation of LLMs — our own language may not be sufficiently adequate to describe how a developmental process should unfold over time in a general and robust way.

As for the final property, using quality-diversity training enables us to forgo having to provide a set of examples for supervised training. Instead, we can define what interesting features characterize relevant entities. However, unlike the approach in \citet{earle2022illuminating} it is not the parameters of the genotype-to-phenotype mapping which we wish to train (i.e.\ the NCAs) but of the DNA input. This will require us to use a slightly different training scheme, which we describe next along with the proposed developmental system.

\subsection{Attention-based guidance}\label{sec:dna-decoding}

Given a genotype-to-phenotype mapping composed of stateful components (such as cells or nodes), we define a guidance mechanism as a function that performs state-conditioned decoding of a shared genome. For a ``DNA''-like genome, we can use a standard attention mechanism \citep{Bahdanau2014NeuralMT} as decoding mechanism as follows. 

Let $\mathbf{h}^i_t$ be the state vectors for at time $t$ for neurons $i$'s and $\mathbf{d}$ be the DNA with characters drawn from the alphabet $\mathcal{A}$. The system first translates each character $d_j$ of $\mathbf{d}$ into a token embedding $\mathbf{e}_j$ (with $j$ being the character position) to create the matrix $\mathbf{E} = [ \mathbf{e}_1, \mathbf{e}_2, ...]$. It then produces a control (or conditioning) vector by applying the attention function\footnote{Notice that we are describing the per-query-operation as opposed to the more standard, but slightly harder to read, batched operation that is usually presented.}:

\begin{equation}
    \mathbf{c}^i_t = v(\mathbf{E})^{\text{T}} \cdot \text{softmax}\Bigg( \frac{\mathbf{E} \cdot q(\mathbf{h}^i_t) }{\sqrt{d}}\Bigg)
\end{equation}
where functions $q: \mathcal{R}^{|h|} \rightarrow \mathcal{R}^{|e|}$ and $v: \mathcal{R}^{|e|} \rightarrow \mathcal{R}^{|h|}$ correspond to the query and value projections (the latter being applied row-wise) whose parameters must be learned by the developmental model, and $d = |\mathbf{\mathbf{d}}|$ is a normalisation constant (see Figure~\ref{fig:model-and-training}a). Notice that we do not require a key projection since, unlike self-attention, the DNA string remains constant throughout the generative process. 

The resulting control vector can then be used to update the internal state of each component:
\begin{equation}
    \mathbf{h}_{t+1}^i = \text{update}(\mathbf{c}^i_t, \mathbf{h}_i^t)
\end{equation}
where the the function $\text{update}(.)$ can be any (learnable) function which produces an output of the same shape as the states. The vector $\mathbf{u}_i^t$ can then be used by downstream operations, enabling the DNA to indirectly guide these operations.

Overall, this mechanism should thus enable two things. First, providing different inputs to the system should change its developmental trajectory, and thus, guide it towards different outcomes. Second, because the decoding depends on the state of each component, these can learn to take on different roles during the unfolding of the developmental process, potentially making said process more expressive and robust. Notice that this directly solves the aforementioned main limitation found in \citet{sudhakaran2021growing}, since the system is not restricted to one-hot encoded goals.

\subsection{Meta-learning using Quality-Diversity algorithms}

Provided with this guidance mechanism we wish to train the developmental mapping (for brevity just mapping) to exploit the potential structure of the DNA genome in order to generate a diverse set of potential outputs. However, unlike more traditional generative models, the system should learn this in an open-ended way,  without having to provide a dataset of examples.

Quality-Diversity (QD) algorithms \citep{pugh2016quality} are a popular class of algorithms, which do not require such explicit supervision while aiming to discover a diverse set of high-quality solutions for a particular problem. In particular, we use the MAP-Elites algorithm \citep{mouret2015illuminating},  which achieves this by combining three components: a measure of the quality of the solutions, a set of behavioral descriptors that define the space of interesting solutions, and an evolutionary algorithm for searching the space. 

At each training iteration an archive containing the best-performing solutions for each discovered descriptor combination is kept. The evolutionary algorithm then generates new potential solutions for which both their descriptor values and quality is determined. If a new descriptor combination is discovered, it is added to the archive, otherwise it replaces the older one if it is of higher quality. Thus the algorithm progressively maximizes the coverage of the space defined by the behavioral descriptors with quality solutions.

Traditionally, this type of approach is applied to the problem of finding a set of parameters (e.g.\ neural network weights) that generate a diverse set of outputs. Here we propose to use these algorithms to evaluate the ability of a genotype-to-phenotype mapping at generating diverse outputs for different \emph{inputs}. 

Given a developmental genotype-to-phenotype mapping $S: \mathcal{D} \rightarrow \mathcal{R}^o$ which generates an output from a genome $\mathbf{d}$, we can evaluate said mapping by computing quality and diversity values for its corresponding output. This enables us to run QD optimization of the encoding space and determine how good the mapping is a exploiting the representation capacity of the genotypes.

More formally, let $\mathbf{A}$ be the archive resulting from a QD algorithm run for a particular mapping $S$, and let $f: \mathcal{R}^o \rightarrow \mathcal{R}$ and $b: \mathbf{R}^o \rightarrow \mathcal{R}^b$ the fitness and descriptor functions respectively. The overall fitness of $S$ is defined as:

\begin{equation}
    \mathcal{s}(S) =  \frac{|\mathbf{A}|} {|\mathbf{A}|_{max}} \cdot  \sum_{\mathbf{d_i} \in \mathbf{A} } f(S(\mathbf{d_i})) 
\end{equation}
where the first term is the coverage achieved: the ratio between the number solutions in the archive ($|\mathbf{A}| = \sum_{\mathbf{d_i} \in \mathbf{A}} 1$) and the maximum number of solutions it could store $|\mathbf{A}|_{max}$, which depend on the binning of the descriptor space $\mathcal{R}^b$. The second term is the accumulated fitness of all the solutions stored.

With this procedure we can evaluate a mapping regardless of implementation. However, if said mapping is a neural network (as in our experiments),  we still need to fit its parameters. To achieve this we can use the above score as the objective function for a meta-training procedure using an evolutionary strategy:

\begin{enumerate}
    \item Initialize an initial population of mappings $S_0$ for the evolutionary algorithm
    \item For a number of training iterations do $n$:
    \begin{enumerate}
        \item Evaluate each $s_i$ with the above procedure.
        \item Adjust the parameters of the evolutionary strategy according to $\mathcal{s}(s_i)$
    \end{enumerate}
    \item After $n$ iterations the set $S_n$ contains the best performing developmental systems.
\end{enumerate}

In other words, the outer loop is tasked with optimizing the developmental system's towards producing as much high quality coverage as possible using the indirect encoding's inherent structure and variability.

\begin{figure*}[t!]
    \centering
    \includegraphics[width=\textwidth]{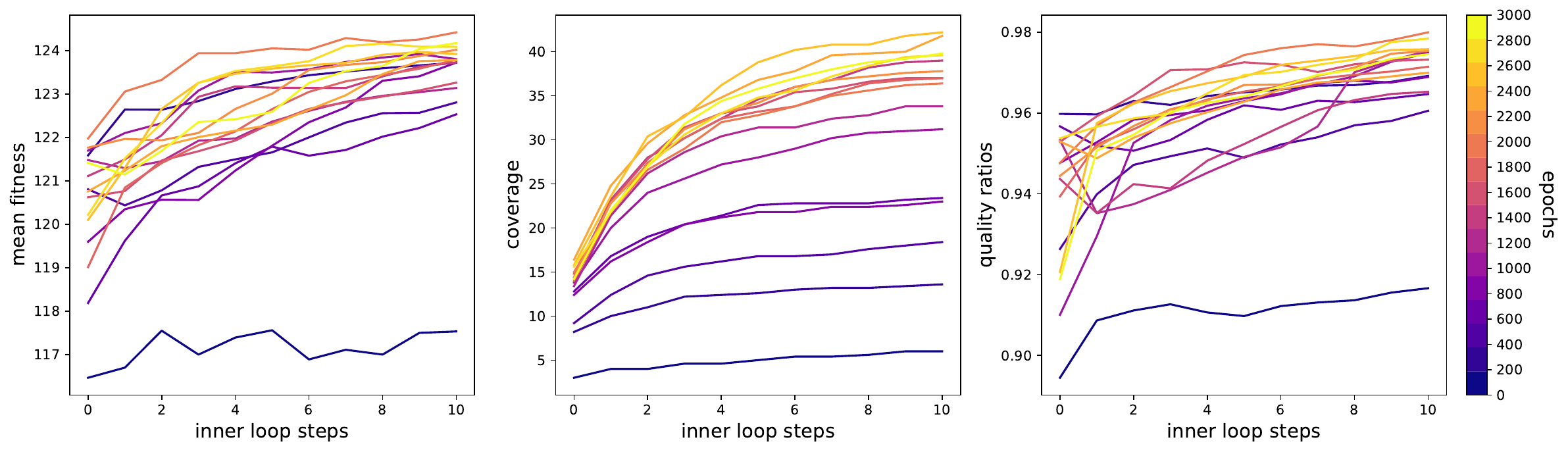}
    \caption{\textbf{Progression of QD metrics across training}. Left panel: the mean fitness scores (y-axis) for solutions in an archive at each step of an inner loop run (x-axis) at different points during training (different colored curves). Middle panel: Same but for the coverage. Right: The quality ratio at each time-step defined as the average quality of the solutions in the archive relative to the maximum possible. These results demonstrate the evolvability of the NCAs. Early in evolution, NCAs produce neither high coverage nor high-quality solutions. After the NCA training (i.e.\ the genotype-to-phenotype mapping), DNA genomes can quickly be evolved that result in high-quality and diverse phenotypes. Note that the first and last panel are similar, but complement each other since the latter tracks the maximum performance that could have been achieved without increasing coverage.
 }
    \label{fig:evolvability}
    \vspace{-0.5cm}
\end{figure*}

An illustration of the overall approach can be seen in Figure~\ref{fig:model-and-training}b. In the next sections we will apply this approach to a standard self-organizing system — a Neural Cellular Automata — to illustrate its ability to generate diverse solutions given a simple input encoding.

\section{Experiments}

We illustrate how the system can be trained with a simple maze generation task. In this task the model must generate mazes  by placing either traversable or wall tiles in a predetermined grid of a particular size. We define the quality function $f(.)$ as the number of connected components in the in the maze (with 1 being the best value). The descriptor function $b(.)$ determines the location of the maze along the axes of maximum minimum-path-length between all pairs of valid tiles and symmetry of the levels (averaged between horizontal and vertical). The grid size is $16 \times 16$ which in turn means that path lengths can go from 0 to $256 / 2 + 16 = 144$ (i.e.\ a zig-zagging pattern across the full width and height of the image, with a cross-section of 1 cell). Symmetry can range from 0 to 128, where 128 equals perfect vertical and horizontal symmetry.

For this problem, the developmental system used is an extension of a Neural Cellular Automata model \citep{mordvintsev2020growing,earle2022illuminating}. The extension is straight-forward following the description in the previous section. That is, we use the state of each cell as queries for the attention-based decoding mechanism. For the DNA encoding, first we set the string size $|\mathbf{s}| = 8$ the alphabet size to $|\mathcal{A}| = 4$. This gives a total of 65,536 possible combinations. Each cell in the NCA has a state size of 9.
In our implementation, the update function is defined as:
\begin{align}
    \mathbf{h}_{t+1}^i &= \mathbf{h}_{t+1}^i + \mathbf{u}_i^t, \\
    \mathbf{u}_i^t &= \text{sign}(\mathbf{z}^i_{s}) \odot \sigma(\mathbf{z}^i_{a}), 
\end{align}
where $[\mathbf{z}^{s}_i; \mathbf{z}^{a}_i] = MLP(\mathbf{c_i^t})$ (i.e.\ the vector is split in half). This non-linearity enables the model to move in any direction of the latent space (by using the sign function) while using the gating to stop the state from being modified if required. We also normalize the resulting hidden state using the standard euclidean norm, which we found improves performance. up This function is a simple extension of the one presented in \cite{springer2020its}, which only uses the sigmoid component and thus limits the state values to only increase in size. 

\begin{figure*}[t!]
    \centering
    \includegraphics[width=\textwidth]{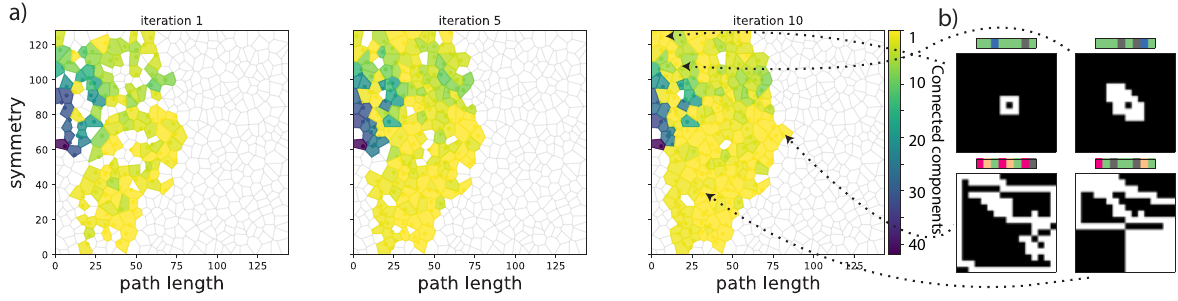}
    \caption{\textbf{Repertoires during inner loop exploration of the encoding space}. a) Each panel shows a different state of the archive during an inner loop run for a trained model. Yellow represents higher-quality solutions and purple low-quality ones as defined by the number of connected components in the maze, with lower being better. b) Example mazes all grown by the same NCA shown together with the associated DNA sequences. The model succeeds at covering almost half of the descriptor space with high-quality solutions. These have different sizes and shapes, showing that the model is effective at controlling its growth. Some similarities between the DNAs can be observed, especially in the top row examples.}
    \label{fig:repertoires}
    \vspace{-0.5cm}
\end{figure*}

Our choice of QD algorithm in the inner loop is the standard MAP-Elites \citep{mouret2015illuminating} with an archive of 500 solutions maximum and a batch size of 100. The algorithm runs for a maximum of 10 iterations, emitting a new batch of solutions using point-wise mutations and uniform cross-over (each of which is used to generate half of the batch). For the former, each character has a $1/|\mathbf{d}|$ chance of being modified, picking one of the other characters with uniform probability $p=1/(|\mathcal{A}| - 1)$. For cross-over, half of the positions in two randomly drawn solutions in the archive are picked uniformly at random and their characters are swapped. For the outer loop, any evolutionary strategy can be applied. In our experiments, we use Covariance Matrix Adaptation (CMA-ES, \citet{hansen2001completely}) with a population size of 100 and keep all other parameters as found in the Evosax implementation \citep{lange2023evosax}. 

All the code for our experiments was implemented in Jax \citep{jax2018github} using Equinox \citep{kidger2021equinox}, Evosax \citep{lange2023evosax} and Qdax \citep{chalumeau2023qdax}. Code to run all the experiments is available \href{https://github.com/miltonllera/meta-evolved-dev}{here}.

\section{Results}

We will first analyse the results for the trained model, and then discuss some important ablations. Figure~\ref{fig:evolvability} shows how the models' evolvability progresses throughout training. We define evolvability in this context as the ability of the model to create high-quality solutions that differ from preexisting high-quality ones. In other words, the solutions added to the archive at each iterations would ideally be as high quality as the pre-existing ones and the model should learn to do this early in the inner loop optimization, i.e.\ it must be easily searchable.

Following this definition, it is easy to see that the system does indeed increase in evolvability as training progresses (as determined by the color of the curves). Not only does the performance at the start of each inner loop become progressively higher further along in training, but so does the rate of change in both total quality and coverage.

Importantly, the model could solve the problem via one of two strategies (or a combination of both): first, compute a set of solutions that achieve maximum coverage, then improve upon them; or it could first obtain a set of solutions of high quality and then mutate them to achieve maximum coverage. The latter is more inline with our notion of evolvability and the panel in Figure~\ref{fig:evolvability} indicates that it is indeed the latter that the system uses. 

In more detail, the results in said panel show the quality ratio — the ratio of average quality for all solutions in the archive at each time-step over the maximum possible value. Thus a value close to 1 for a particular iteration indicates that the model first optimizes and then covers the space. This effect also appears early in training, with the model adopting said strategy within the first 200 outer loop iterations. This happens in spite of the fact that coverage and  has a multiplicative effect on the total performance $f$, which would favor the opposite strategy. Note however that this does not by itself imply that the model is able to generalise — i.e. generate solutions far outside the space that it tends to cover during training. We will discuss possible implications of this issue in the final section.

Furthermore, in Figure~\ref{fig:repertoires} we visualize the repertoires themselves as the inner loop progresses, showing three different archives for different steps in said inner loop of a trained model. We can see that the model does indeed cover a large section of the space in only 10 iterations. For comparison, \citet{earle2022illuminating} trains the model on a similar problem for $10000$ iterations on a $3000$ sized parameters space vs only $10$ and $8$ in our solution (albeit achieving higher coverage and not requiring a meta loop). Additionally, we observe some structure in the corresponding DNA strings.

\begin{figure*}[!t]
    \centering
    \includegraphics[width=\textwidth]{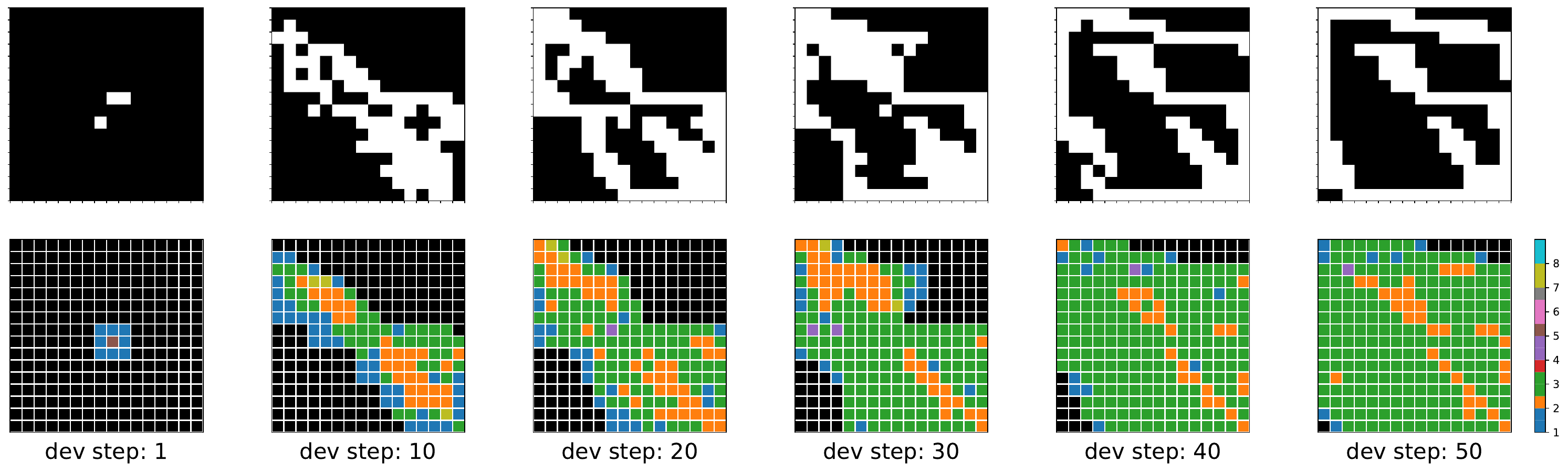}
    \caption{\textbf{Growing mazes using DNA-guided NCAs}. Each column in the top row shows a different step in the models development of a maze, from step 1 to 50. The bottom row shows the position of the DNA sequence with the highest attention weight for each cell in the NCA at the corresponding time step. Dark cells represent cells that are not yet alive. The attention maps show that these follow a similar pattern as the resulting shape, pointing to how each position is likely being used by the system to grow the corresponding shape.}
    \label{fig:level-gen}
    \vspace{-0.5cm}
\end{figure*}

Figure~\ref{fig:level-gen} on the other hand shows several steps of a trained NCA while generating a level. In the top row we see, how starting from the center of the grid, the model slowly builds a pattern of valid tiles. In this case, the model achieves perfect quality, so as growth progresses it must decide to convert cells into valid tiles in order to maintain connectivity.

In the bottom row, we see how the model attends to different positions in the DNA as development progresses. This provides a simple visualization of how the model uses the DNA in potentially interesting ways. Notice that cells at the edge of the generated map (adjacent to black tiles) predominantly attend to the first position of the DNA as they become alive, but can switch to predominantly attending to other positions. On the other hand, cells in the central diagonal of the pattern stay constant, attending to the second position in the DNA, which suggests that they use this as a reference for which pattern to construct.

\subsection{Analysing the system}

How does the developmental system use the genotype space to achieve its goal? Does any structure emerge as the result of training? In other words, are specific positions or characters in the DNA representation associated with specific properties of the output?

Testing whether latent representations in neural networks exhibit structure is an active area of research in the Deep Learning community. For this work, we will make use of three metrics from research on the topic of disentanglement \citep{eastwood2018framework} and emergent languages \citep{chaabouni2020compositionality}.

Specifically, we can measure the total disentanglement of the model by measuring how each position in the DNA is predictive of each behavioral feature. This procedure is based on fitting a Random Forest Regression model \citep{breiman2001random} for each dependent variable and measuring the feature importance of each input for that prediction. Highly disentangled features will only be predictive of one dependent variable (or a small subset of them). The overall level of disentanglement of a model is then computed as the average disentanglement over all features.
In the study of emergent languages, a set of related metrics have been proposed by \citet{chaabouni2020compositionality} based on the ``information gap'' of a feature with respect to the predicted properties. Briefly, the information gap of a feature is the difference in mutual information between it and two properties witch which it has highest mutual information. The higher the information gap the more predictive the feature is one property and not the rest. We use two measures called position information gap and character information gap. These account for how predictive a character at a particular position and the number of repetitions of a character in the DNA string are informative of the properties of the generated mazes.

\begin{figure}[b!]
    \centering
    \includegraphics[width=0.95\linewidth]{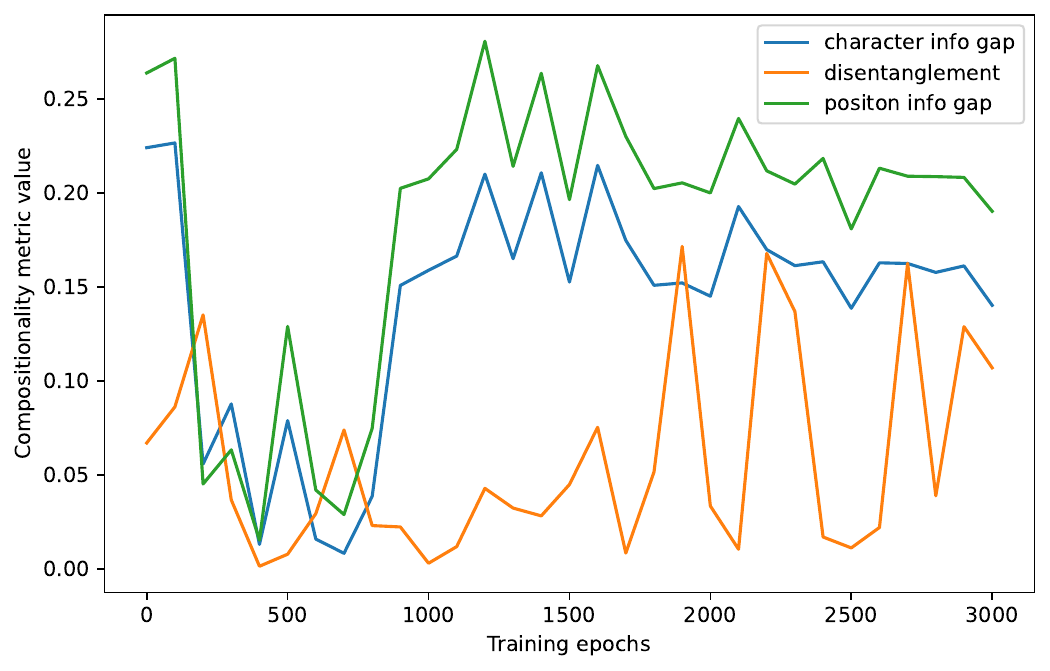}
    \caption{\textbf{Measuring the structure of the DNA representation}. The values for three metrics which measure potential structure that appears in DNA-space as training of the developmental system progresses. This indicates that there is emergent structure in the genotype space as a result of training, but that the model is likely susceptible to becoming chaotic as this can be an effective strategy for covering the behavioral space.}
    \label{fig:compositionality}
    \vspace{-0.3cm}
\end{figure}

The results of this analysis can be observed in Figure~\ref{fig:compositionality}. We see that there is an initial structure present in the model, likely do to random initialisation and the nature of the attention mechanism (which tends to amplify the most attended position). This quickly disappears until roughly a third of the way through training when the measures that compute emergence of language-like features (position info gap and character info gap) show an increase in their value. However, disentanglement remains low until later in training, when inf starts to fluctuate rapidly. At this point the other two metrics dip slightly. This indicates that the model may have a propensity towards over-fitting and/or using a chaotic strategy relying more on its inherent stochasticity then on the structure of the DNA space. Such a limitation would have to be addressed in future approaches.

\subsection{Controlling for different design decisions}

Having analyzed the system we now test the effect of different design decisions. First, the most important test we must perform is controlling for the effect of using a purely continuous genotype (as a opposed to DNA one). Second, we must test how modifying the update function affects model changes model performance, since this is the critical component that enables self-organisation. Specifically, we will show that normalisation of the states after update is performed significantly affects model performance. Finally, we will control for the usage of a more sophisticated emitter in the inner loop based on the CMA evolutionary strategy instead of the standard GA.

\begin{figure}[t!]
    \centering
    \includegraphics[width=\linewidth]{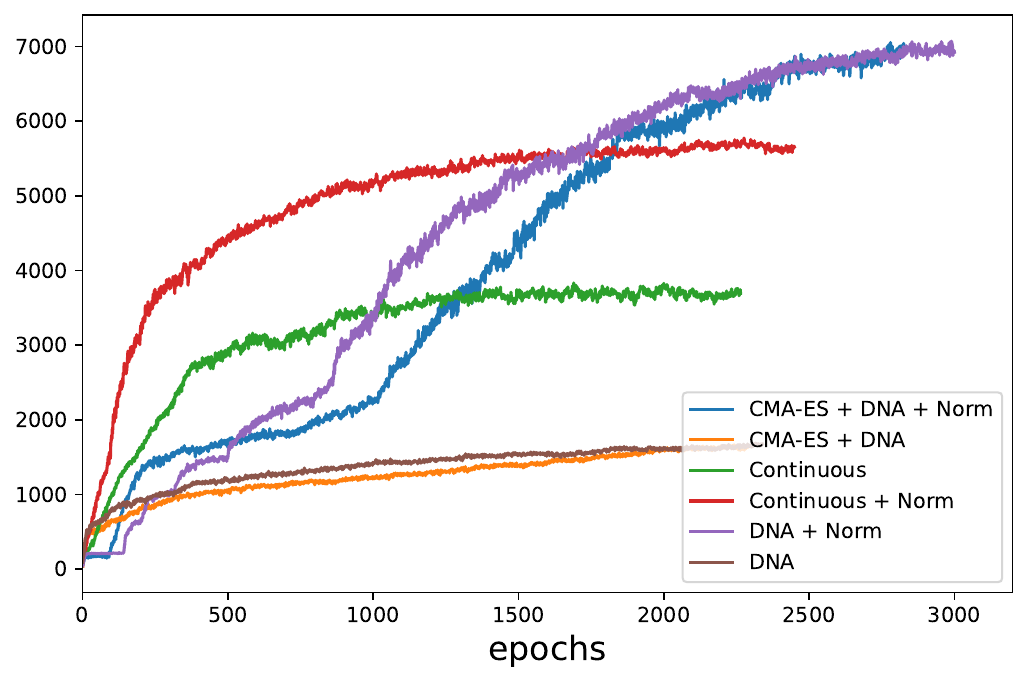}
    \caption{\textbf{Controlling for key design decisions}. The plot shows the training performance by QD score of several runs where different components were swapped out. \textcolor{violet}{Violet} and \textcolor{blue}{blue} curves show the performance of a system using DNA genotypes, one using GA and the other using CMA in the inner loop. \textcolor{brown}{Brown} and \textcolor{orange}{orange} are the same, but the states of the model are not normalized after updating. \textcolor{red}{Red} and \textcolor{green}{green} are models using continuous vectors serving as genotypes with and without normalization after updates, respectively. Some decisions like the choice of inner loop emitter seem to have less of an impact on performance, while others like normalization matter more. Importantly, using a continuous vector facilitates fast convergence, but is ultimately less expressive than a DNA genotype.}
    \label{fig:controls}
    \vspace{-0.5cm}
\end{figure}

Figure~\ref{fig:controls} shows the results for 6 combinations of controls: DNA with and without normalization (violet, brown), DNA trained with CMA with and without normalization (blue, orange) and continuous vector genotype, again with and without normalization (red, green). These results show that the system using a DNA genotype performs better than the one using a continuous vector in spite of the latter converging quicker. However, this is only true if we normalize the states of the model after the update, indicating once again that the system is more susceptible to instability. Finally, using CMA produces no significant difference in performance. This is rather surprising considering that this algorithm is widely used in this context due to its efficiency and shows that the DNA encoding is powerful enough to enable efficient exploration without a complex evolutionary algorithm.

\section{Discussion}

The separation between the DNA and the developmental mechanisms that transcribe the information contained in it, underpin the evolvable nature of biological systems. Harnessing such capabilities, even if only minimally, could boost the capabilities of AI systems, enabling more autonomous and robust systems that could enact complex behaviour. Our results show that it is indeed possible to integrate these principles with an artificial developmental system. This enables it to exploit information in genotypic space to produce a diverse set of phenotypes. 

Indeed, as training progresses the NCA learns to exploit variability in the DNA space to increase it's coverage of the descriptor space with high quality solutions. The fact that the model tends to prefer maximizing the quality of it's solutions before increasing coverage shows that it has some evolvable properties. Importantly, we can analyse how the DNA space is used during development. This is possible because this architecture makes explicit the parts of the DNA that are being attended to throughout development, which makes it easier to understand its dynamics.

Nonetheless, there are still obvious limitations. First, the system doesn't exhibit strong evolvability — i.e. if we extend the number of inner loops iterations, the model does not substantially increase its coverage of the descriptor space. Thus, the model is still fundamentally limited in its ability to effectively generalise to out-of-distribution outputs. Such property is however fundamental to bridging the gap with biological entities and to enable try open-ended innovation.

Second, the system shows a tendency towards structural properties in the DNA space degrading over time. Thus as training progresses we observe structure metrics decay (see Figure~\ref{fig:compositionality}). The fact that the model can still generate consistent structures may indicate that it tends to over-fit on particular patterns that increase coverage with minimal modifications. Thus the DNA space does not need to be very structured for the model to succeed.

Future research will be focused on addressing these issues. In our view, the causes of these limitations are twofold. First, while the system is in principle very expressive, this may come at the cost of inductive biases that favor the reuse of basic patterns in different combinations and at different scales. Second, it may be the case that the current descriptors allow the model to ``cheat'' by exploiting the fact that minimal differences can crate diversity (for example only removing one tile can increase the minimum path length significantly) without requiring fundamental changes to the pattern being generated. Future research would thus explore two directions: different developmental system architectures and different task descriptions. Of particular interest would be to combine this approach with Neural Developmental Programs for Neural Architecture Search \citep{najarro2023towards,nisioti2024}.

\section{Acknowledgements}
This project was supported by a European Research Council (ERC) grant (GA no. 101045094, project ”GROW-AI”).

\footnotesize
\bibliographystyle{apalike}
\bibliography{bibliography} 

\end{document}